\documentclass[runningheads]{llncs}

\usepackage{graphicx}
\usepackage{amsmath,amssymb} 
\usepackage{color}
\usepackage{hyperref}
\usepackage{algorithm}
\usepackage[noend]{algpseudocode}
\DeclareMathOperator{\E}{\mathbb{E}}

\newcommand{\etal}{\textit{et al}.}

\begin{document}

\title{Learning to Anonymize Faces for \\Privacy Preserving Action Detection} 
\titlerunning{Privacy Preserving Action Detection}
\author{Zhongzheng Ren\inst{1,2}\orcidID{0000-0003-1033-5341}
\and Yong Jae Lee\inst{1,2}\orcidID{0000-0001-9863-1270}
\and Michael S. Ryoo\inst{1}\orcidID{0000-0002-5452-8332}}
%


\institute{EgoVid Inc., South Korea
\and University of California, Davis \\ 
\email{\{zzren,yongjaelee\}@ucdavis.edu, mryoo@egovid.com}}

\authorrunning{Z. Ren, Y. J. Lee and M. S. Ryoo}

\maketitle
\begin{abstract}
There is an increasing concern in computer vision devices invading users' privacy by recording unwanted videos. On the one hand, we want the camera systems to recognize important events and assist human daily lives by understanding its videos, but on the other hand we want to ensure that they do not intrude people's privacy. In this paper, we propose a new principled approach for learning a video \emph{face anonymizer}. We use an adversarial training setting in which two competing systems fight: (1) a video anonymizer that modifies the original video to remove privacy-sensitive information while still trying to maximize spatial action detection performance, and (2) a discriminator that tries to extract privacy-sensitive information from the anonymized videos. The end result is a video anonymizer that performs pixel-level modifications to anonymize each person's face, with minimal effect on action detection performance. We experimentally confirm the benefits of our approach compared to conventional hand-crafted anonymization methods including masking, blurring, and noise adding. Code, demo, and more results can be found on our project page \url{https://jason718.github.io/project/privacy/main.html}.

\end{abstract}

\section{Introduction}
\label{sec:intro}

Computer vision technology is enabling automatic understanding of large-scale visual data and is becoming a crucial component of many societal applications with ubiquitous cameras. For instance, cities are adopting networked camera systems for policing and intelligent resource allocation, individuals are recording their lives using wearable devices, and service robots at homes and public places are becoming increasingly popular. 



\begin{figure*}[t!]
\centering
\includegraphics[width=1\textwidth]{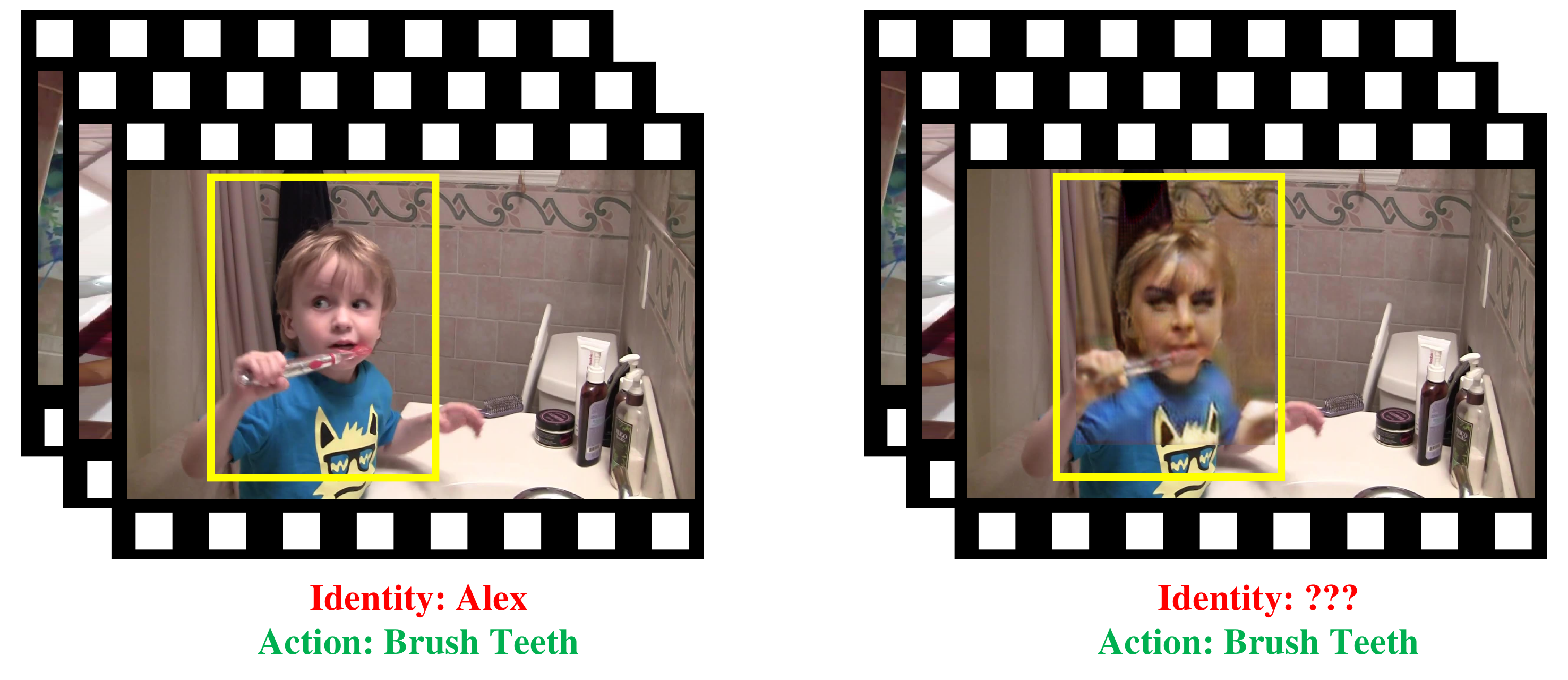}
\caption{Imagine the following scenario: you would like a personal assistant that can alert you when your adorable child Alex performs undesirable actions, such as eating mom's make-up or drinking dirty water out of curiosity.  However, you do not want your personal assistant to record Alex's face, because you are concerned about his privacy information since the camera could potentially be hacked.  Ideally, we would like a face anonymizer that can preserve Alex's privacy (i.e., make his face no longer recognizable as Alex) while at the same time unaltering his actions.  In this paper, our goal is to create such a system. (Real experimental results.)}
\label{figure:concept}
\end{figure*}


Simultaneously, there is an increasing concern in these systems invading the privacy of their users; in particular, from unwanted video recording. On the one hand, we want the camera systems to recognize important events and assist human daily lives by understanding its videos, but on the other hand we want to ensure that they do not intrude the people's privacy. Most computer vision algorithms require loading high resolution images/videos that can contain privacy-sensitive data to CPU/GPU memory to enable visual recognition. They sometimes even require network access to high computing power servers, sending potentially privacy-sensitive images/videos. All these create a potential risk of one's private visual data being snatched by someone else. In the worst case, the users are under the risk of being monitored by a hacker if their cameras/robots at home are cracked. There can also be hidden backdoors installed by the manufacturer, guaranteeing their access to cameras at one's home.

To address these concerns, we need a principled way to `anonymize' one's videos. Existing anonymization methods include extreme downsampling~\cite{ryoo-aaa17} or image masking, as well as more advanced image processing techniques using image segmentation~\cite{butler15}. Although such techniques remove scene details in the images/videos in an attempt to protect privacy, they are based on heuristics rather than being learned, and there is no guarantee that they are optimal for privacy-protection. Moreover, they can hurt the ensuing visual recognition performance due to loss of information~\cite{ryoo-aaa17}. Thus, a key challenge is creating an approach that can simultaneously anonymize videos, while ensuring that the anonymization does not negatively affect recognition performance; see Fig.~\ref{figure:concept}.


In this paper, we propose a novel principled approach for \emph{learning} the video anonymizer. We use an adversarial training strategy; i.e., we model the learning process as a fight between two competing systems: (1) a video anonymizer that modifies the original video to remove privacy-sensitive information while preserving scene understanding performance, and (2) a discriminator that extracts privacy-sensitive information from such anonymized videos. We use human face identity as the representative private information---since face is one of the strongest cues to infer a person's identity---and use action detection as the representative scene understanding task.


To implement our idea, we use a multi-task extension of the generative adversarial network (GAN)~\cite{gan} formulation. Our face anonymizer serves as the generator and modifies the face pixels in video frames to minimize face identification accuracy.  Our face identifier serves as the discriminator, and tries to maximize face identification accuracy in spite of the modifications. The activity detection model serves as another component to favor modifications that lead to maximal activity detection. We experimentally confirm the benefits of our approach for privacy-preserving action detection on the DALY~\cite{daly} and JHMDB~\cite{jhmdb} datasets compared to conventional hand-crafted anonymization methods including masking, blurring, and noise adding.



Finally, although outside the scope of this work, the idea is that once we have the learned anonymizer, we could apply it to various applications including surveillance, smart-home cameras, and robots, by designing an embedded chipset responsible for the anonymization at the hardware-level. This would allow the images/videos to lose the identity information before they get loaded to the processors or sent to the network for recognition.

\section{Related work}
\label{sec:related}

\paragraph{\textbf{Privacy-Preserving Recognition}}
There are very few research studies on human action recognition that preserve identity information. The objective is to remove the identity information of people appearing in `testing' videos (which is a bit different from protecting privacy of people in training data~\cite{differential_privacy16,yonetani17}), while still enabling reliable recognition from such identity-removed videos.

Ryoo \etal ~\cite{ryoo-aaa17} worked on learning of efficient low resolution video transforms to classify actions from extreme low resolution videos. Chen \etal ~\cite{chen17} extended \cite{wang16} for low resolution videos, designing a two-stream version of it. Ryoo \etal ~\cite{ryoo-aaa18} further studied the method of learning a better representation space for such very low resolution (e.g., 16x12) videos. All these previous works relied on video downsampling techniques which are hand-crafted and thus not guaranteed to be optimal for privacy-preserving action recognition. Indeed, the low resolution recognition performances of these works were much lower than the state-of-the-art on high resolution videos, particularly for large-scale video datasets.
Jourabloo \etal~\cite{Jourabloo} de-identify faces while preserving facial attributes by fusing faces with similar attributes. It achieves impressive results on gray-scale facial pictures with attribute annotations.  However, it is specific to the facial attribute setting and is not applicable to more general privacy-sensitive tasks such as action recognition.

\paragraph{\textbf{Action Detection}}

Action recognition has a long research history~\cite{aggarwal11}. In the past several years, CNN models have obtained particularly successful results. This includes two-stream CNNs \cite{simonyan2014two,feichtenhofer2016convolutional} and 3-D XYT convolutional models \cite{tran2014c3d,carreira2017quo} for action classification, as well as models for temporal action detection \cite{sigurdsson2016hollywood,piergiovanni2018super}.

Our paper is related to spatial action detection from videos, which is the task of localizing actions (with bounding boxes) in each video frame and categorizing them. Recent state-of-the-art spatial action detectors are modified from object detection CNNs models. Gkioxari and Malik~\cite{actiontubes} extend the R-CNN~\cite{rcnn} framework to a two-stream variant which takes RGB and flow as input.  Weinzaepfel \etal~\cite{weinzaepfel-2015} improved this method by introducing tracking-by-detection to get temporal results.  Two-stream Faster R-CNN~\cite{ren16faster} was then introduced by \cite{saha-2016,peng2016}.  Singh \etal~\cite{singh-2017} modified the SSD~\cite{ssd} detector to achieve real-time action localization.  In this work, we use Faster RCNN~\cite{ren16faster} as our frame-level action detector.

\paragraph{\textbf{Face Recognition}}
Face recognition is a well-studied problem~\cite{facesurvey1,facesurvey2}. Recent deep learning methods and large-scale annotated datasets have dramatically increased the performance on this task~\cite{10000face,deepface,centerloss,facenet,Liu_2016_large,Liu_2017_CVPR}. Some approaches treat it as a multi-class classification problem and use a vanilla softmax function~\cite{10000face,deepface}.  Wen \etal~\cite{centerloss} introduce the ``center loss'' and combine it with the softmax loss to jointly minimize intra-class feature distance while maximizing inter-class distance. The state-of-the-art approach of~\cite{facenet} uses the triplet loss with hard instance mining but it requires 200 millions training images. Recent work~\cite{Liu_2016_large,Liu_2017_CVPR} demonstrate strong performance by combining metric learning with classification. In this work, we use~\cite{Liu_2017_CVPR} as our face recognition module. 

\paragraph{\textbf{Network Attacking}}
Our work is also closely related to the problem of network attacking. Existing CNN based classifiers are easily fooled~\cite{junyan-iclr18,carlini17,Goodfellow-2015,Moosavi-Dezfooli17,PapernotMG16} even when the input images are perturbed in an unnoticeable way to human eyes. Correspondingly, there is also a line of work studying defense methods~\cite{Goodfellow-2015,PapernotM0JS16,XuEQ17}. Our work is similar to network attacking methods since our modified images need to attack a face identifier. However, the difference is that we want to dramatically change the content, so that the identity is unrecognizable (even for humans), while also optimizing it for action recognition.

\paragraph{\textbf{Generative Adversarial Networks (GANs)}}
GANs~\cite{gan} have been proposed to generate realistic images in an unsupervised manner. Since then, numerous works~\cite{wgan,SalimansGZCRCC16,dcgan} have studied ways to improve training of GANs to generate high-quality and high-resolution images. It is currently the most dominant generative model and the key to its success is the adversarial loss, which forces the generated data distribution to be indistinguishable from the real one.  GANs have been generalized and applied to various vision tasks such as image-to-image translation~\cite{pix2pix}, super resolution~\cite{ledigTHCATTWS16}, domain adaptation~\cite{ren-cvpr2018}, and object detection~\cite{a-fast-rcnn}. Recent work uses GANs to suppress user information in video features for privacy~\cite{iwasawa17}, but it only focused on feature extraction without considering modification of actual pixels in image data. In this paper, we extend the GAN formulation to learn an explicit \emph{face modifier} to anonymize each person's face without hurting action detection performance. Also, compared to image-to-image translation, style transfer, and domain adaptation works, our network does not require a target domain to borrow the visual style or content from.

\section{Approach}
\label{sec:approach}

\begin{figure*}[t!]
\centering
\includegraphics[width=\textwidth]{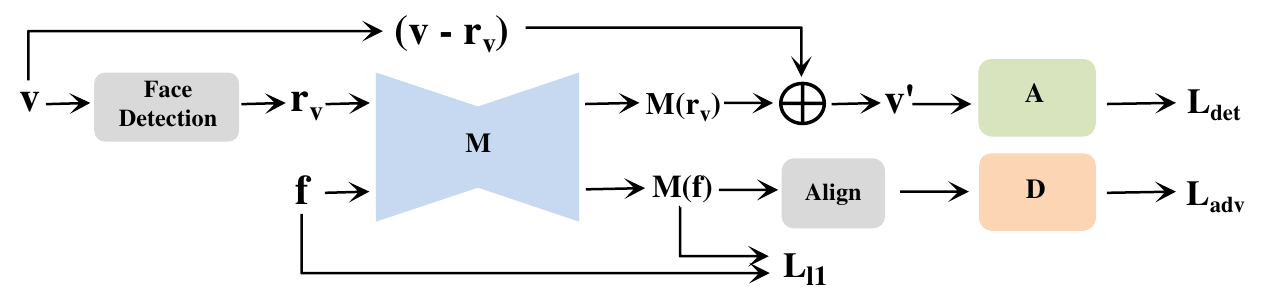}
\caption{Our network architecture for privacy-preserving action detection.  We simultaneously train a face modifier $M$ whose job is to alter the input face ($f$ or $r_v$) so that its identity no longer matches that of the true identity, and an action detector $A$ whose job is to learn to accurately detect actions in videos in spite of the modifications.  The face classifier $D$ acts as an adversary and ensures that the modified face is non-trivial.  See text for details. (Gray blobs are not learned during training.)}
\label{figure:archi}
\end{figure*}

Given a set of training videos $V$ and face images $F$, our goal is to learn a face modifier that anonymizes each person in the video frames and images (i.e., so that they cannot be correctly recognized by a face recognition system) while simultaneously training an action detector that can detect each person's action in spite of the modification.  We formulate the problem as a multi-task learning objective.  The overall framework is depicted in Fig.~\ref{figure:archi}.  There are three major learnable components: the modifier $M$ which takes a face image as input and anonymizes it, an action detector $A$ which detects the actions of the people in each video frame, and a face classifier $D$ that classifies the identity of each face.  

There are two main advantages in using both videos and images for training. First, we can leverage existing large-scale labeled face datasets to learn identity information. Second, we can train our model on action detection datasets that do not have any identity annotations.  In other words, we do not need to create a new specific dataset to train our model, but instead can leverage the (disparate) datasets that were created for face recognition and action detection without any additional annotations.

We next introduce the loss functions and then explain the training procedure. The implementation details are given at the end.

\subsection{Formulation}

Our loss for training the model consists of three parts: an adversarial classification loss for identity modification; an action detection loss for training the action detector; and an L1 loss to encourage each generated image to preserve as much structure (pose, brightness, etc.) of the original unmodified face as possible.

\paragraph{\textbf{Action Detection Loss}}
Given an input frame from video dataset $v \in V$, we first apply a face detector to get face region $r_v$. We then feed $r_v$ into the modifier $M$ and replace $r_v$ with the resulting modified image $M(r_v)$ in the original frame to get $v' = v-r_v+M(r_v)$. In other words, the frame is identical to the original frame except that the face of the person has been modified.  (If there are multiple people in the frame, we repeat the process for each detected person's face.)

The per-frame action detector is trained using the modified frame $v'$, and ground-truth action bounding boxes $\{b_i(v)\}$ and corresponding action category labels $\{t_i(v)\}$.  Specifically, the detection loss is defined as:
\begin{equation}
    L_{det}(M,A,V) = \E_{v \sim V} [L_{A}(v',\{b_i(v)\},\{t_i(v)\})]
\label{eq:det}
\end{equation}
where $L_{A}$ is the sum of the four losses in Faster-RCNN~\cite{ren-cvpr2018}: RPN classification and regression, and Fast-RCNN classification and regression.  We choose Faster-RCNN as it is one of the state-of-the-art object detection frameworks that has previously been used for spatial action detection successfully (e.g.,~\cite{saha-2016,peng2016}).

\paragraph{\textbf{Adversarial Classification}}
Using a state-of-the-art face classifier, we can easily achieve high face verification accuracy~\cite{Liu_2016_large,Liu_2017_CVPR}. In particular, we use the face classifier formulation of~\cite{Liu_2017_CVPR} to be the target discriminator for our setting. In order to fool it, our modifier $M$ needs to generate a very different-looking person. Simultaneously, the face classifier $D$ should continuously be optimized with respect to the anonymized faces $M(f)$, so that it can correctly identify the face despite any modifications. Our $D$ is initialized with pre-trained parameters learned from large-scale face datasets.

We use an adversarial loss to model this two-player game~\cite{gan}.  Specifically, during training, we take turns updating $M$ and $D$.  Here, we denote the input image from the face dataset as $f \in F$ and the corresponding identity label as $i_f \in I$.  The loss is expressed as:
\begin{equation}
L_{adv}(M, D, F) = - \E_{(f \sim F, i_f \sim I)}[L_{D}(M(f), i_f)] - \E_{(f \sim F, i_f \sim I)}[L_{D}(f, i_f)].
\label{eq:adv}
\end{equation}
Here the classification loss $L_{D}$ is the angular softmax loss~\cite{Liu_2017_CVPR}, which has been shown to be better than vanilla softmax via its incorporation of a metric learning objective. When updating $M$ this loss is minimized, while when updating $D$ it is maximized.  This simultaneously enforces a good modifier to be learned that can fool the face classifier (i.e., make it classify the modified face with the wrong identity), while the face classifier also becomes more robust in dealing with the modifications to correctly classify the face despite the modifications. Furthermore, we optimize $D$ for face classification using both the modified images $M(f)$ and the original images $f$. We find that this leads to the modifier producing faces that look very different from the original faces, as we show in the experiments.




\paragraph{\textbf{Photorealistic Loss}}
We use an L1 loss to encourage the modified image to preserve the basic structure (pose, brightness, etc.) of the original picture.  The L1 loss was previously used in image translation work~\cite{pix2pix,CycleGAN} to force visual similarity between a generated image and the input image.  Although this loss does not directly contribute to our goal of privacy-preserving action detection, we add it because we want the modified image to retain enough scene/action information that can also be recognizable by a human observer. At the same time since we want to ensure enough modification so that the person's identity is no longer recognizable, we use a weight $\lambda$ and set its value to be relatively small to avoid making the modified image look too similar to the original one:
\begin{equation}
    L_{l1}(M,F) = \E_{f \sim F} [~\lambda~||M(f) - f||_{1} ]
\label{eq:l1}
\end{equation}

\paragraph{\textbf{Full Objective}}
Our full objective is:
\begin{equation}
L(M,D,A,V,F)= L_{det}(M,A,V) + L_{adv}(M, D, F) + L_{l1}(M, F) 
\label{eq:all}
\end{equation}
We aim to solve:
\begin{equation}
    \operatorname*{arg}\min_{M, A} \max_D ~L(M,D,A,V,F)
\end{equation}
Our algorithm's pseudo code for learning is shown in Alg.~\ref{alg:1}. There are two things to note: (1) if there are no frontal faces in a frame (due to occlusion or if the person is facing away from the camera), we use the original, unmodified frame to train the action detector; (2) During training, we update the face classifier, modifier, and action detector iteratively. Therefore, the input image to the action detector keeps changing, which can make its optimization difficult and unstable. To overcome this, once the loss terms for the modifier and face classifier converge, we fix the modifier and face classifier and only fine-tune the action detector. A similar training procedure is used for our baseline approaches in Section 5, with the only difference being the modification procedure.


\begin{algorithm}[!t]
\footnotesize
\caption{Privacy Preserving Action Detection} \label{trainalg}
    \textbf{Input:} Video frames $V$ and action labels; Face images $F$ and identity labels; Face classifier $D$; Training iteration $T_1, T_2$\\
    \textbf{Output:} Face modifier $M$; Privacy preserving action detector $A$
\begin{algorithmic}[1]
    \For{$t = 1$ to ${T_1}$}
      	\State $M(f)\rightarrow f'$ \hfill // Face modification
      	\State $\operatorname*{arg\,max}_D~~L_{adv}(M,D,F) $\hfill // Update $D$
      	\State $det\_face(v) \rightarrow r_v $ \hfill // Face detection
      	\State if $\#faces\_in\_frame > 0$   \hfill // Video frame modification
      	\State ~~~~~~~$M(r_v)\rightarrow r_v',  (v-r_v)+r_v' \rightarrow v'$ 
      	\State else    \hfill // No Faces to modify
      	 \State ~~~~~~~$v \rightarrow v'$  
      	\State $\operatorname*{arg\,min}_{M,A}~~L_{det}(M,A,V) + L_{adv}(M, D, F) + \lambda~L_{l1}(M, F) $ \hfill  // Update $M, A$
    \EndFor
    \For{$\tau = 1$ to ${T_2}$}    
      	\State $\operatorname*{arg\,min}_{A}L_{det}(M,A,V)$  \hfill  // Freeze $M,D$; Update $A$
    \EndFor
\end{algorithmic}
\label{alg:1}
\end{algorithm}

\section{Implementation}
\paragraph{\textbf{Face Detection}}
We use SSH~\cite{ssh} face detector to detect faces in our video dataset, which produces high recall but with some false positives. Therefore, we keep the detections with probability greater than 0.8 and feed the rest into the MTCNN~\cite{mtcnn} face detector to remove false positives by using it as a binary classifier. After these two steps, we get a clean and highly-accurate set of face bounding boxes.

\paragraph{\textbf{Face Modification}}
We adapt the architecture for our modifier from Johnson \etal~\cite{Johnson2016Perceptual}, which has demonstrated impressive image translation performance on various datasets by Zhu \etal~\cite{CycleGAN}. We use the 9 residual blocks network and instance normalization~\cite{inst_norm}. The input images are upsampled or sampled to $256\times 256$ via bilinear interpolation.

\paragraph{\textbf{Spatial Action Detection}}
We use Faster-RCNN~\cite{ren16faster} with ResNet-101~\cite{resnet} as the backbone network for action detection and train it end-to-end. Images are resized so that the shorter length is 340 pixels for JHMDB and 600$\times$800 pixels for DALY following Gu \etal~\cite{ava}.

\paragraph{\textbf{Face Identity Classification}}
We use the SphereFace-20~\cite{Liu_2017_CVPR} network, which combines metric learning and classification to learn a discriminative identity classifier. We use the CASIA-WebFace pretrained~\cite{casia} model for initialization. The input face images are aligned and cropped using facial keypoints. We use a differentiable non-parametric grid generator and sampler for cropping (similar to the warping procedure in Spatial Transformer Networks~\cite{stn} except there are no learnable parameters) to make our whole network end-to-end trainable.

\paragraph{\textbf{Training Details}}

We use the Adam solver~\cite{adam}, with momentum parameters $\beta_1= 0.5$, $\beta_2=0.999$. A learning rate of 0.001 for Faster RCNN, 0.0003 for face modifier and face classifier. We train the entire network for 12 epochs and drop the learning rate by $\frac{1}{10}$ after the seventh epoch.

\section{Results}
\label{sec:results}
In this section, we first provide details on the evaluation metrics and datasets, and then explain the baseline methods. We then evaluate our method's performance both quantitatively and qualitatively. In addition, we conduct a user study to verify whether the modified photos can fool human subjects. Finally, we perform ablation studies to dissect the contribution of each model component.

\subsection{Metrics and Datasets}
\paragraph{\textbf{Action Detection}}
We measure detection performance using the standard mean Average Precision (mAP) metric. The IoU threshold $\delta$ is set to 0.5 when measuring spatial localization. We use two datasets: DALY~\cite{daly} and JHMDB (split1)~\cite{jhmdb}, as they contain a number of actions that involve the face area and thus are good testbeds for our joint face anonymization and action detection model. For example, some of the action classes have similar body movement and hand motions near the head (taking photos, phoning, brushing hair) or mouth (drinking, brushing teeth, playing harmonica, applying make up on lips).

\paragraph{\textbf{Face Recognition}}
Following previous works~\cite{Liu_2016_large,Liu_2017_CVPR}, we measure face recognition performance by training our model on face classification and evaluating on binary face verification. CASIA-WebFace~\cite{casia} is used for training and the input images are aligned and cropped using the facial keypoints estimated by MTCNN~\cite{mtcnn}. During testing, we extract the features after the last fully-connected layer and then compute cosine similarity for face verification on LFW \cite{LFWTech}, one of the most popular face datasets for this task. Note that here our motivation is not to create a new face recognition model but rather to use an established method as an adversary for our adversarial setting.

\begin{figure*}[t!]
\centering
\includegraphics[width=\textwidth]{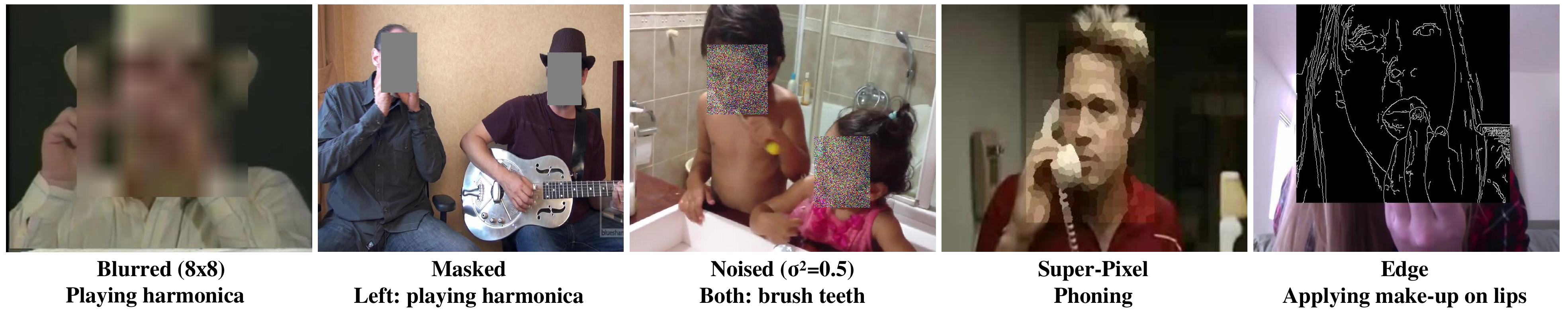}
\caption{Baseline modification examples. Although these methods largely conceal the true identity, they can also be detrimental to action detection performance especially if the action involves the facial region. (Zoom in for details, better viewed in pdf.)}
\label{figure:modi}
\end{figure*}

\subsection{Baselines}
One straightforward and brute force solution to address privacy concerns is to detect faces and modify them using simple image processing methods such as blurring, masking, and additive noise, etc. To explore whether they are viable solutions, we use several of them as our baselines and evaluate their performance on both action detection and face recognition. For action detection, the detected face boxes are enlarged by $1.6\times$ to ensure that they include most of the head region, and then are modified. This enlargement also helps the video face region $r_v$ be more similar to the face image $f$, which has some background context (see examples in Fig.~\ref{figure:quali1} top).

We want to ensure that the baseline face anonymization methods are strong enough to preserve privacy (i.e., make the original face identity unrecognizable to humans). With this motivation, we implemented the following methods: (1) \textbf{Un-anonymized}: no protection is applied; (2) \textbf{Blur}: following Ryoo~\etal~\cite{ryoo-aaa17}, we downsample the face region to extreme low-resolution ($8\times8, 16\times16, 24\times24$) and then upsample back; (3) \textbf{Masked}: the faces are masked out; (4) \textbf{Noise}: strong Gaussian noise is added ($\sigma^2=0.1,0.3,0.5$); (5) \textbf{Super-pixel}: following~\cite{butler15}, we use superpixels and replace each pixel's RGB value with its superpixel's mean RGB value; (6) \textbf{Edge}: following~\cite{butler15}, face regions are replaced by their corresponding edge map. Example modifications are shown in Fig.~\ref{figure:modi}.

\begin{figure}[t!]
  \begin{minipage}{0.5\linewidth}
    \centering
    \includegraphics[scale = 0.4]{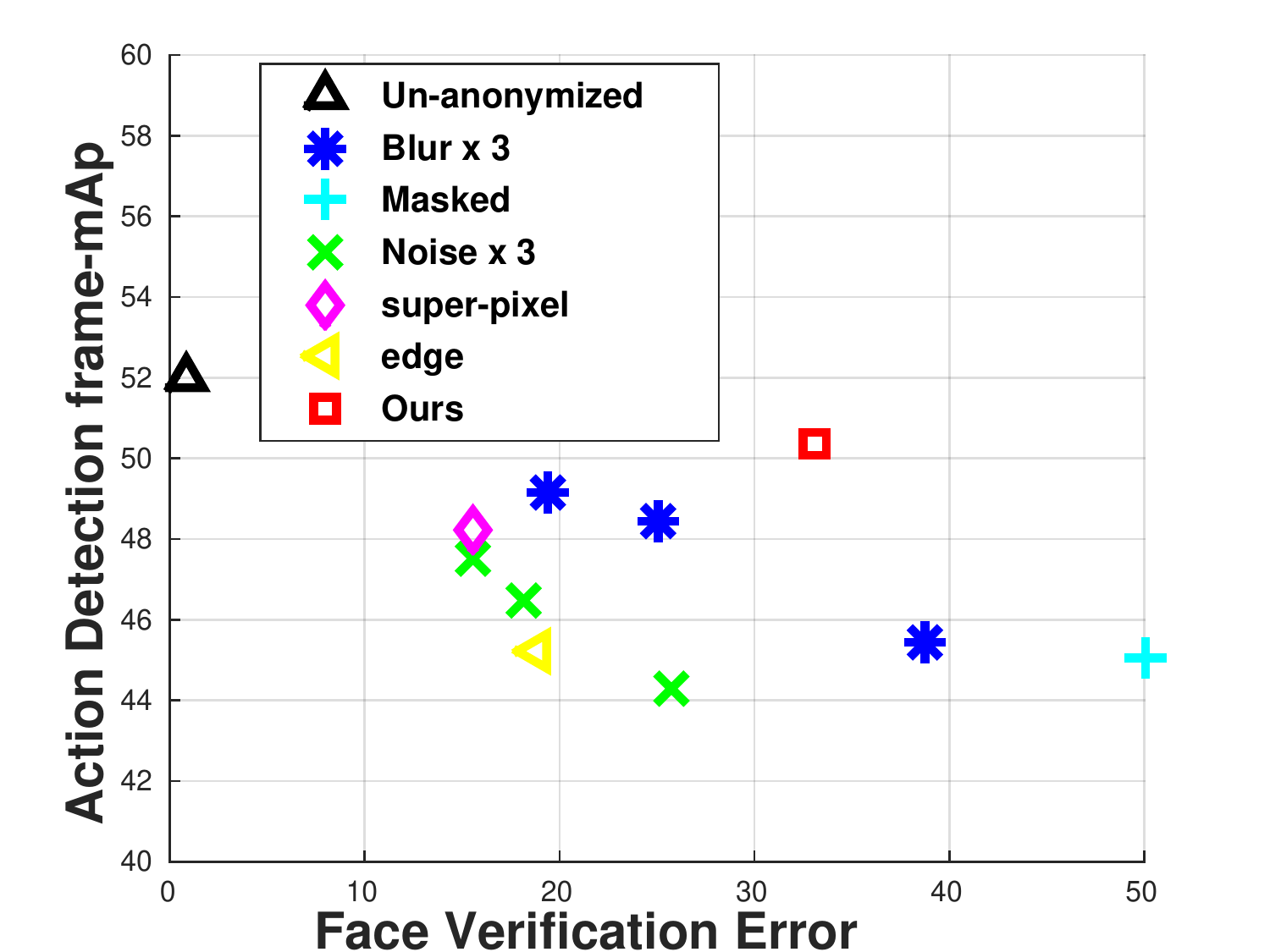}
  \end{minipage}%
  \begin{minipage}{0.5\linewidth}
    \centering
    \includegraphics[scale = 0.4]{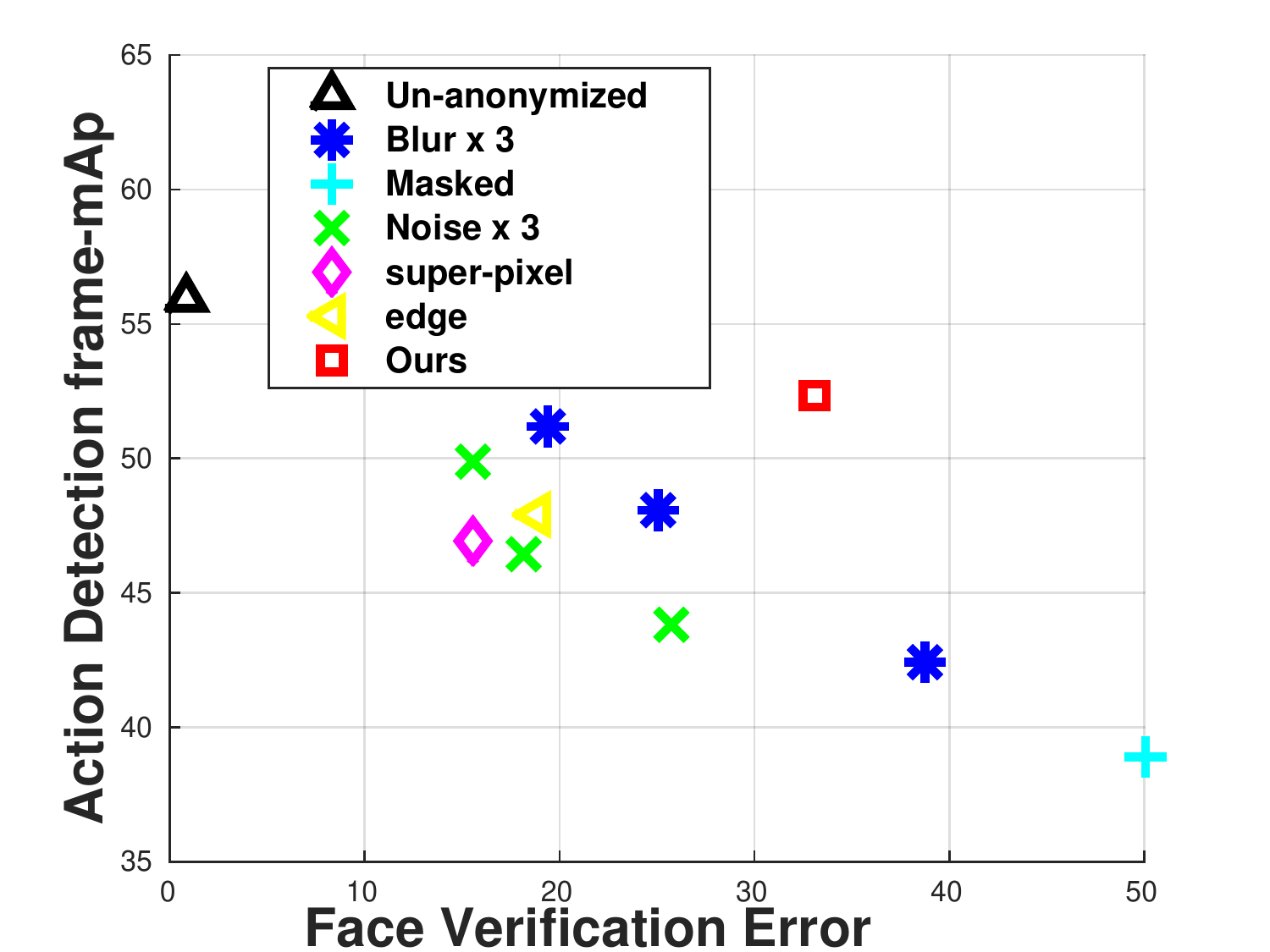}
  \end{minipage}
\caption{x-axis is face verification error and y-axis is action detection mAP. The closer a model's performance is to the top-right corner, the better. \textbf{Left:} JHMDB; \textbf{Right:} DALY.}
\label{figure:res}
\end{figure}

Fig.~\ref{figure:res} shows action detection accuracy (y-axis) vs.~face verification error (x-axis) for the baselines, with JHMDB results on left and DALY results on right. As expected, the baselines greatly increase face recognition error (and thus improving anonymization) compared to the original un-anonymized frames. However, at the same time, they also harm action detection performance on both DALY and JHMDB. These results indicate that simple image processing methods are double-edged swords; although they may be able to protect privacy (to varying degrees), the protection comes with the negative consequence of poor action recognition performance.

\subsection{Quantitative Results}
\paragraph{\textbf{Overall Performance}}
Fig.~\ref{figure:res} also shows ours results indicated by the \textbf{red square} maker.  Our method simultaneously optimizes on both tasks and achieves better results compared to the baselines.  As two extreme one-sided solutions, the `Un-anonymized' (top-left) and `Masked' (bottom-right) baselines can only address either action detection or face verification.   Our action detection results are significantly better than others while being quite close to the un-anonymized detection results. For face verification, our method is only worse than two baselines ($8\times8$ down-sampling and masking) but outperforms the others.

\begin{table}[t!]
\scriptsize
\centering
\begin{tabular}{c | c c  c  c  c  c  c  c  c  c | c }
\hline
Action   & Lip & Brush & Floor & Window & Drink & Fold & Iron & Phone & Harmonica & Photo & mAP\\
\hline
Un-anonymized  & 92.30 & 51.26 & 76.73	& 27.87 & 31.23& 32.67 & 75.30 & 51.50 & 73.91 & 55.74 & 56.85 \\
\hline
Blur(8x8)     & 84.31 & 17.87 & 79.39 & 27.77 & 6.60  & 28.68 & 71.69 & 28.40 & 31.13 & 48.09 & 42.39 \\
Blur(16x16)   & 82.07 & 32.40 & 79.53 & 32.14 & 10.74 & 31.35 & 74.91 & 36.97 & 48.37 & 52.51 & 48.10\\
Blur(24x24)   & \textbf{92.08} & \textbf{39.84} & 79.93 & 31.77 & 15.23 & 34.96 & 74.65 & 46.33 & 51.78 & 53.09 &51.97\\
Noise($\sigma^2=0.1$)   
              & 87.71 & 31.37 & 78.41 & 31.87 & 12.41 & 34.80 & 76.50 & 42.37 & 50.14 & 53.48 & 49.91\\
Noise($\sigma^2=0.3$)   
			  & 87.64 & 24.98 & 78.59 & \textbf{32.68} & 8.34  & 35.33 & 74.96 & 40.12 & 36.61 & 45.42 & 46.47\\
Noise($\sigma^2=0.5$)
			 & 83.63 & 21.45 & \textbf{81.32} & 29.59 & 7.43  & 29.08 & 77.97 & 33.93 & 27.35 & 46.35 & 43.81\\
Masked 	     & 67.06 & 15.19 & 78.86 & 26.58 & 6.59  & 25.53 & 72.95 & 27.79 & 21.32 & 46.76 &38.86 \\
Edge         & 80.30 & 29.46 & 78.02 & 30.51 & 10.31 & 32.64 & \textbf{79.15} & 35.15 & 54.69 & 49.62 & 47.99\\
Super-Pixel  & 79.47 & 26.09 & 80.82 & 32.22 & 11.46 & \textbf{35.29} & 77.70 & 30.30 & 42.18 & 53.68 & 46.92\\
\hline
Ours    & 89.20 & 33.08 & 77.12 & 32.56 & \textbf{22.93} & 33.86 & 77.07 & \textbf{46.52} & \textbf{55.32} & \textbf{55.54} & \textbf{52.32}\\
\hline
\end{tabular}
\caption{Action detection accuracy on DALY. `Lip', `Brush', `Floor', `Window', `Drink', `Fold', `Iron', `Phone', `Harmonica', `Photo', denote category `applying make-up on lips', `brushing teeth', `cleaning floor', `cleaning windows', `drinking', `folding textile', `ironing', `phoning', `playing harmonica', `taking photos or videos'.}
\label{tbl:class_DALY}
\end{table}

\paragraph{\textbf{Per Class Accuracy}}
As described earlier, there are certain actions that are more influenced by face modification. Therefore, we next investigate per-class detection results to closely analyze such cases. As shown in Table~\ref{tbl:class_DALY}, we find that our model boosts action detection accuracy with a bigger margin compared to the baselines if the actions are `drinking', `phoning', `playing harmonica', and `taking photos or videos'. This result makes sense since these actions involve the face areas and our approach best preserves the original pictures' semantics.  Our model only marginally improves over or performs worse for `cleaning floor', `cleaning windows', `ironing' because these actions have almost nothing to do with faces. Overall, these results indicate that our model modifies each face in a way that ensures high action detection performance.

\subsection{Qualitative Results}

\begin{figure}[t!]
\centering
\includegraphics[width=1\textwidth]{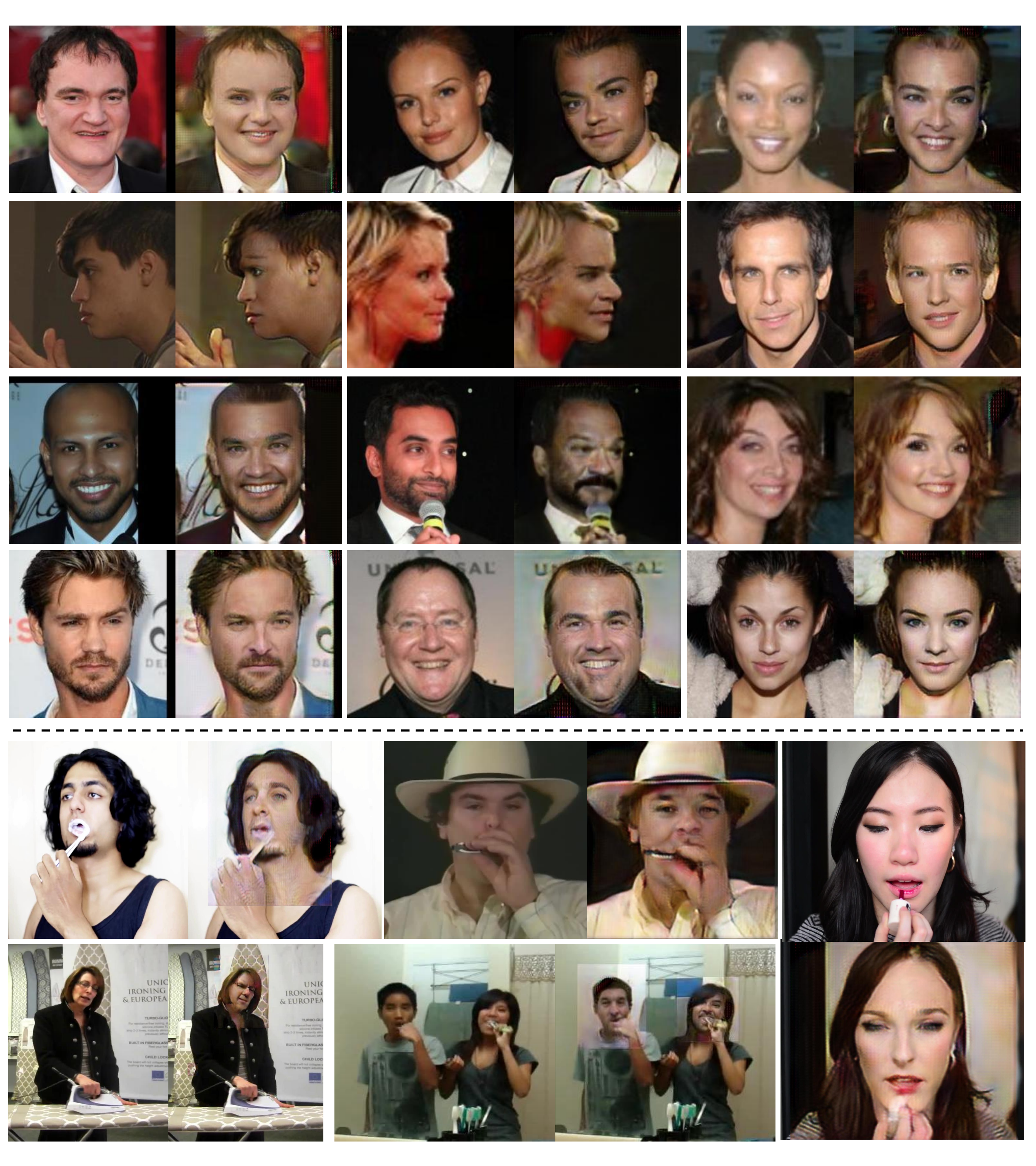}
\caption{The same image before and after anonymization. The picture on the left of each pair is the original image, and the one on the right is the modified image. The first four rows are from the face dataset; the bottom two are from the video dataset.} 
\label{figure:quali1}
\end{figure}

\begin{figure}[t!]
\centering
\includegraphics[width=1\textwidth]{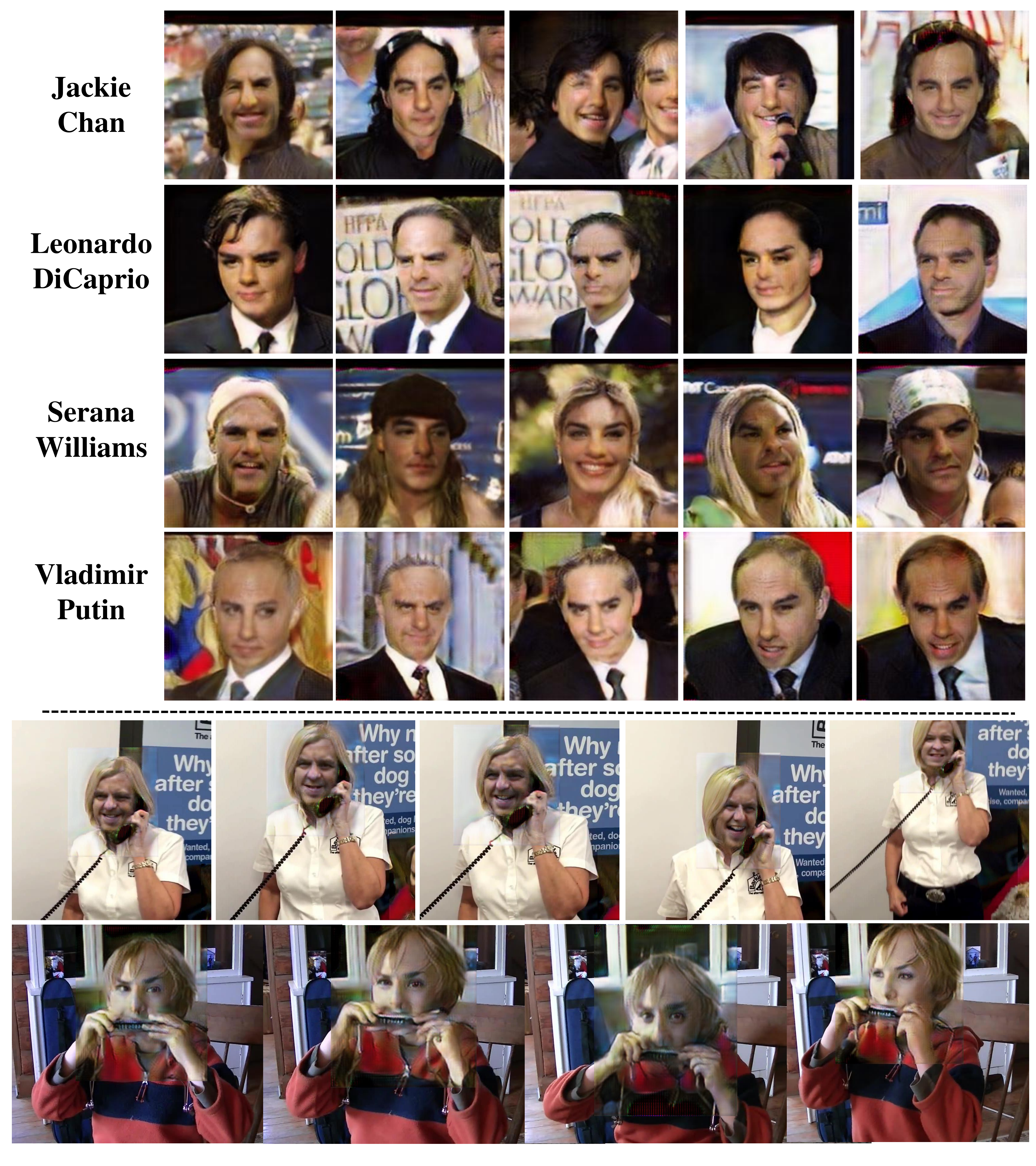}
\caption{Anonymized images. Top: different modified pictures of the same person; Bottom: different modified frames in the same video.} 
\label{figure:quali2}
\end{figure}

\paragraph{\textbf{Same picture before and after modification}} We next show examples of pictures before and after modification in Fig.~\ref{figure:quali1}. The first four rows show images from the face dataset, while the bottom two rows show images from the video dataset.  Overall, we can see that our modifier generates realistic pictures that change the person's identity. Importantly, the basic structure (pose, clothing, background) and the actions (brushing teeth, playing harmonica, putting makeup on lips) are preserved, showing the contribution of the $L1$ and action detection losses.

To change the person's identity, our network tends to focus more on local details. For example, in the first row, our model changes the gender; in the third row, the hair style (baldness and color) is changed; in the fourth row, facial details like nose, eye glasses, and eyebrow get modified. We can make the same observation for the video frame results: the two teeth brushing teenagers become older; the ethnicity of the woman who is putting makeup on her lips changes. 

\paragraph{\textbf{Different modified pictures of the same person}} Here we explore how our model modifies different face images of the same person.  This is to answer whether our model first recognizes the person and then systematically changes his/her identity, or whether it changes the identity in a more stochastic manner.   

Fig.~\ref{figure:quali2} shows the results. The set of original images (prior to modification; not shown here) in each row all have the same identity.  The first four rows show modified images of: Jackie Chan, Leonardo DiCaprio, Serena Williams, and Vladimir Putin.  The bottom two rows show different modified frames of the same person in the same video.  Given the consistency in the modifications for the same person (i.e., the modified faces look similar), it appears that our model is recognizing the identity of the person and systematically modifying it to generate a new identity.  This result is surprising because there is nothing in our objective that forces the model to do this; one hypothesis is that this happens because we are using gradient ascent to maximize face classification error and it is an easier optimization to reshape the face identity manifold in a systematic way compared to perturbing each face instance separately. 


\subsection{User study}
We conducted a very simple user study to investigate how well our modifier can fool humans. We designed three questions: (Q1) We sample a pair of modified images from the testing set and ask our subjects whether the pair corresponds to the same person. We collect 12 positive pairs and 12 negative pairs. (Q2) We use our model to modify 16 famous celebrities in LFW (who are not in our training data) and ask our subjects to name them or say `no idea'. (Q3) We display a set of modified images, and ask our subjects if they think this technology is good enough to protect their own privacy.

In total, we collected 400 answers from 10 different subjects. The overall accuracy for Q1 is $53.3\%$, which is close to random guessing ($50\%$). For Q2, among the celebrities they know, our subjects could only name $19.75\%$ of them correctly based on the modified images. Finally for Q3, all subjects except 2 responded that they felt their identity information would be protected if they were to use this technology. 


\subsection{Ablation Study}
\paragraph{\textbf{Does the face classifier remain accurate?}}
During the training process, we observed that the generator (i.e., modifier) tends to ``win'' over the discriminator (i.e., face classifier).  This raises the concern that our face classifier may no longer be robust enough to correctly classify an unmodified face.  To answer this, we take the trained face classifier and evaluate it on the original un-anonymized LFW faces.  The classifier's features produce $94.75\%$ verification accuracy.  On modified LFW faces, they only achieve $66.95\%$. This shows that the classifier can still accurately recognize the original faces despite being ``fooled'' by the modified faces.   


\paragraph{\textbf{Gradient ascent or use random label when optimizing M?}}
Inspired by existing network attacking works~\cite{junyan-iclr18,carlini17,Goodfellow-2015,Moosavi-Dezfooli17,PapernotMG16}, we can also optimize our modifier so that it fools the classifier to classify the modified image as a random face category. (In our approach, as shown in Alg.~\ref{alg:1}, we instead perform gradient ascent to maximize classification loss.) In practice, we find that random negative sample optimization produces much worse results where the resulting generated faces have obvious artifacts and lose too much detail.

One possible explanation for this is that the optimization for this baseline is much harder compared to gradient ascent (i.e., maximizing  classification error for the correct identity). Here, the optimization target keeps changing randomly during training, which leads to the entire network suffering from mode collapse.  Thus, it simply produces a consistent blur regardless of the original identity. In contrast, gradient ascent makes the modified image still look like a face, only with a different identity.


\section{Conclusion}
We presented a novel approach to learn a face anonymizer and activity detector using an adversarial learning formulation. Our experiments quantitatively and qualitatively demonstrate that the learned anonymizer confuses both humans and machines in face identification while producing reliable action detection.

\paragraph{Acknowledgements}
This research was conducted as a part of EgoVid Inc.'s research activity on privacy-preserving computer vision, and was supported in part by the Technology development Program (S2557960) funded by the Ministry of SMEs and Startups (MSS, Korea), and NSF IIS-1748387.  We thank all the subjects who participated in our user study. We also thank Chongruo Wu, Fanyi Xiao, Krishna Kumar Singh, and Maheen Rashid for their valuable discussions. 

\bibliographystyle{splncs04}
\bibliography{egbib}

\end{document}